\documentclass[twoside]{article} 

\usepackage[accepted]{aistats2017}
\usepackage{amsmath}
\usepackage{amssymb}
\usepackage{graphicx}
\usepackage{booktabs}
\usepackage[acronym,xindy,smallcaps]{glossaries}
\usepackage{glossaries}

\makeglossaries
\newacronym[longplural={Gaussian Processes}]{gp}{gp}{Gaussian Process}
\begin{document}

%

%

\twocolumn[

\aistatstitle{Distribution of Gaussian Process Arc Lengths}

\aistatsauthor{Justin D. Bewsher* \And Alessandra Tosi \And Michael A. Osborne \And Stephen J. Roberts }

\aistatsaddress{University of Oxford \And Mind Foundry, Oxford \And University of Oxford \And University of Oxford} ]

\begin{abstract}
  
  We present the first treatment of the arc length of the \gls{gp} with more than a single output dimension. \Glspl{gp} are commonly used for tasks such as trajectory modelling, where path length is a crucial quantity of interest. Previously, only paths in one dimension have been considered, with no theoretical consideration of higher dimensional problems.  We fill the gap in the existing literature by deriving the moments of the arc length for a stationary \gls{gp} with multiple output dimensions. A new method is used to derive the mean of a one-dimensional \gls{gp} over a finite interval, by considering the distribution of the arc length integrand. This technique is used to derive an approximate distribution over the arc length of a vector valued \gls{gp} in $\mathbb{R}^n$ by moment matching the distribution. Numerical simulations confirm our theoretical derivations. 
\end{abstract}

\section{INTRODUCTION}

\glsreset{gp}\Glspl{gp} \cite{Rasmussen2004} are a ubiquitous tool in machine learning. They provide a flexible non-parametric approach to non-linear data modelling. \Glspl{gp} have been used in a number of machine learning problems, including latent variable modelling \cite{Lawrence2005}, dynamical time-series modelling \cite{Wang2005} and Bayesian optimisation \cite{Snoek2012}.

At present, a gap exists in the literature; we fill that gap by providing the first analysis of the moments of the arc length of a vector-valued \gls{gp}. Previous work tackles only the univariate case, making it inapplicable to important applications in, for example, medical vision (or brain imaging) \cite{Hauberg2015} and path planning \cite{Moll2004}. The authors believe that an understanding of the arc length properties of a \gls{gp} will open up promising avenues of research. Arc length statistics have been used to analyze multivariate time series modelling \cite{Wickramarachchi2015}. In \cite{Tosi2014}, the authors minimize the arc length of a deterministic curve which then implicitly defines a \gls{gp}. In another related paper \cite{Hennig2014a}, the authors use \glspl{gp} as approximations to geodesics and compute arc lengths using a na\"ive Monte Carlo method.

We envision the arc length as a cost function in Bayesian optimisation, as a tool in path planning problems \cite{marchant2014bayesian} and a way to construct meaningful features from functional data. 

Consider a Euclidean space $X = \mathbb{R}^n$ and a differentiable injective function $\gamma:[0,T] \to \mathbb{R}^n$. Then the image of the curve, $\gamma$, is a curve with length:
\begin{equation}
\text{length}(\gamma) = \int^{T}_{0}|\gamma'(t)|\mathrm{d}t.
\end{equation}
Importantly, the length of the curve is independent of the choice of parametrization of the curve \cite{docarmo:1992}. For the specific case where $X = \mathbb{R}^2$, with the parametrization in terms of $t$, $\gamma = (y(t), x(t))$, we have:
\begin{equation}
\text{length}(\gamma) = \int^{T}_{0}|\gamma'(t)|\mathrm{d}t = \int^{T}_{0}\sqrt{y'(t)^2 + x'(t)^2}\mathrm{d}t.
\end{equation}
If we can write $y = f(x)$, $x=t$, then our expression reduces to the commonly known expression for the arc length of a function:
\begin{equation}
\text{length}(\gamma) = s = \int^{T}_{0}|\gamma'(t)|\mathrm{d}t = \int^{T}_{0}\sqrt{1 + \left(f'(t)\right)^2}\mathrm{d}t,
\end{equation}
where we have introduced $s$ as a shorthand for the length of our curve. In some cases the exact form of $s$ can be computed, for more complicated curves, such as Beizer and splines curves, we must appeal to numerical methods to compute the length.


Our interest lies in considering the length of a function modelled with a \gls{gp}. Intuition suggests that the length of a \gls{gp} will concentrate around the mean function with the statistical properties dictated by the choice of kernel and the corresponding hyperparameters.

Previous work \cite{BARAKAT1970} considered the derivative process $f'(t)$, which is itself a \gls{gp}. A direct calculation was performed and an exact form for the mean was obtained in terms of modified Bessel functions; a form for the variance is also presented. This result is also derived in \cite{Miller1956,Corrsin1961}. Analysis has been presented on the arc length of a high-level excursion from the mean \cite{NOSKO1985}. 

However, other important questions have not been explored within the literature. In particular, the shape of the distribution has not been communicated, computing the arc length of a posterior \glspl{gp} has not been addressed, nor has anyone considered the arc length of \glspl{gp} in anything other than $\mathbb{R}$. These issues are addressed within this paper; we present a new derivation of the mean of a one dimensional \gls{gp} and derive the moments of a \gls{gp} in $\mathbb{R}^n$.

The paper is structured as follows. In Section \ref{sec:theory} we review the theory of \glspl{gp} and introduce the notation necessary to deal with \gls{gp}s defined on $\mathbb{R}^n$. In Section \ref{sec:one_d} we examine the one dimensional case, deriving a distribution over the arc length increment before computing the mean and variance of the arc length. In Section \ref{sec:general_case} we consider the general case. A closed form distribution is not possible for the increment, therefore we provide a moment-matched approximation that proves high-fidelity to the true distribution. This distribution allows us to compute the corresponding moments for the arc length. Section \ref{sec:numerical} presents numerical simulations demonstrating the theoretical results. Finally we conclude with thoughts on the use of arc length priors.

\section{GAUSSIAN PROCESSES}
\label{sec:theory}
\subsection{Single Output Gaussian Processes}

Consider a stochastic process from a domain $f: \mathcal{X} \to \mathbb{R}$. Then if $f$ is a \gls{gp}, with mean function $\mu$ and kernel $k$, we write
\begin{align}
 f \sim \mathrm{GP}(\mu,k).
\end{align}
We can think of a \gls{gp} as an extension of the multivariate Gaussian distribution for function values and as the multivariate case, a \gls{gp} is completely specified by its mean and covariance function. 
For a detailed introduction see \cite{Rasmussen2004}. 
Given a set of observations $S = \{x_i, y_i\}_{i=1}^{N}$ with Gaussian noise $\sigma^2$ the posterior distribution for an unseen data, $x_{*}$, is
\begin{align}
 p(f(x_{*})|S,x_{*},\phi) & = \mathcal{N}(f(x_*),m(x_*),k(x_{*},x_{*})).
\end{align}
Letting $X = [x_1, \dots, x_n]$, and defining $k_{x_{*}} = K(X,x_{*})$, the posterior mean and covariance are:
\begin{align}
 m(x_{*}) & = k_{x_*}^T(k(X,X) + \sigma^2\mathbf{I})^{-1}y\\
 C_{*}(x_{*},x_{*}) & = k(x_{*},x_{*}) - k_{x_*}^T(k(X,X) + \sigma^2\mathbf{I})^{-1}k_{x_{*}}
\end{align}
The derivative of the posterior mean can be calculated:
\begin{align}
    \frac{\partial m_*}{\partial x_*} & = \frac{\partial k(x_*,X)}{\partial x_{*}}(k(X,X)+ \sigma^2\mathbf{I})^{-1}y,\label{eqn:post_mean_deriv}
\end{align}
wherever the derivative of the kernel function can be calculated. The covariance for the derivative process can likewise be derived and hence a full distribution over the derivative process can be specified.

Alternatively, a derivative \gls{gp} process can be defined for any twice-differentiable kernel in terms of our prior distribution. If $f \sim \mathrm{GP}(\mu, k)$, then we write the derivative process as
\begin{align}
  f' \sim \mathrm{GP}(\partial \mu,\partial^2k).
\end{align}
The auto-correlation, $\rho(\tau)$, of the derivative process is related to the autocorrelation of the original process via the relation \cite{Middleton1960}:
\begin{align}
 \rho_{f'}(\tau) & = -\frac{\mathrm{d}^2}{\mathrm{d}\tau^2}\rho_{f}(\tau) = \frac{\partial^2}{\partial x\partial x'}k(x-x'),
\end{align}
where $\tau = x-x'$. The variance of $f'$ is therefore given by $\sigma_{f'}^2 = \rho_{f'}(0)$.

\subsection{Vector Valued Gaussian Processes}

The development of the multi-output \glspl{gp} proceeds in a manner similar to the single output case; for a detailed review see \cite{Alvarez2012}. 
The outputs are random variables associated with different processes evaluated at potentially different values of $\mathbf{x}$. 
We consider a vector valued \gls{gp}:
\begin{align}
 \mathbf{f} & \sim \mathrm{GP}(\mathbf{m}, \mathbf{K}),
\end{align}
where $\mathbf{m} \in \mathbb{R}^D$ is the mean vector where $\{m_d(x) \}_{d=1}^{D}$ are mean functions associated with each output and $\mathbf{K}$ is now a positive definite matrix valued function. $(\mathbf{K}(\mathbf{x},\mathbf{x}'))_{d,d'}$ is the covariance between $f_d(x)$ and $f_{d'}(x')$. Given input $\mathbf{X}$, our prior over $\mathbf{f}(\mathbf{X})$ is now
\begin{align}
    \mathbf{f}(\mathbf{X}) \sim \mathcal{N}(\mathbf{m}(\mathbf{X}), \mathbf{K}(\mathbf{X},\mathbf{X})).
\end{align}
$\mathbf{m}(\mathbf{X})$ is a $DN$-length vector that concatenates the mean vectors for each output and $\mathbf{K}(\mathbf{X},\mathbf{X})$ is a $ND \times ND$ block partitioned matrix. In the vector valued case the predictive equations for an unseen datum, $\mathbf{x_{*}}$ become:
\begin{align}
 \mathbf{m}(\mathbf{x}_{*}) & = \mathbf{K}_{\mathbf{x}_*}^T(\mathbf{K}(\mathbf{X},\mathbf{X}) + \mathbf{\Sigma})^{-1}\mathbf{y}\\
 \mathbf{C}_{*}(\mathbf{x}_{*},\mathbf{x}_{*}) & = \mathbf{K}(\mathbf{x}_{*},\mathbf{x}_{*}) - \mathbf{K}_{x_*}^T(\mathbf{K}(\mathbf{X},\mathbf{X}) + \mathbf{\Sigma})^{-1}\mathbf{K}_{\mathbf{x}_{*}},
\end{align}
where $\mathbf{\Sigma}$ is block diagonal matrix with the prior noise of each output along the diagonal. The problem now focuses on specifying the form of the covariance matrix $\mathbf{K}$. We are interested in separable kernels of the form:
\begin{align}
 \mathbf{K}(\mathbf{x},\mathbf{x}')_{d,d'} & = k(\mathbf{x},\mathbf{x}')k_T(d,d'),
\end{align}
where $k$ and $k_T$ are themselves valid kernels. The kernel can then be specified in the form:
\begin{align}
 \mathbf{K}(\mathbf{x},\mathbf{x}') & = \mathbf{k}(\mathbf{x},\mathbf{x}')\mathrm{B} 
\end{align}
where $\mathrm{B}$ is a $D \times D$ matrix. For a data set $\mathbf{X}$:
\begin{align}
 \mathbf{K}(\mathbf{X},\mathbf{X}) & = \mathrm{B}\otimes k(\mathbf{X},\mathbf{X}),
\end{align}
with $\otimes$ representing the Kronecker product. $\mathrm{B}$ specifies the degree of correlation between the outputs. Various choices of $\mathrm{B}$ result in what is known as the Intrinsic Model of Coregionalisation (IMC) or Linear Model of Coregionalisation (LMC).


\subsection{Kernel Choices}

A \gls{gp} prior is specified by a choice of kernel, which encodes our belief about the nature of our function behaviour. In the case of infinitely differentiable functions we might choose the exponentiated quadratic, for periodic functions, the periodic kernel, or for cases where we wish to control the differentiability of our function we might select the Mat\'{e}rn class of kernels \cite{Rasmussen2004}. Samples from each kernel result in distinct sample curve behaviour. 

We demonstrate how the choice of kernel impacts the statistical behaviour of our arc length and relate this to the kernel hyperparameters of several popular kernels. For the vector-valued \gls{gp} we show the statistical properties are related to the choice of the spatial kernel coupled with the choice of the output dependency matrix $\mathrm{B}$ and in particular, its eigenvalues. This highlights how kernel choice affects not only the shape but the length of our functions or conversely how knowledge of the prior curve length could be used to inform kernel selection.








\section{ONE DIMENSIONAL ARC LENGTH}
\label{sec:one_d}
First we consider the one dimensional case, where we develop a new method to derive the expected length; an approach that can be used in the vector case. Consider a \gls{gp}, $f \sim \mathrm{GP}(0,\mathrm{K})$ with a corresponding derivative process $f' \sim \mathrm{GP}(0, \partial^2\mathrm{K})$. Then the arc length is the quantity:
\begin{align}
 s & = \int_{a}^{b}\sqrt{1 + (f')^2}\mathrm{d}t \label{eqn:oned_length}.
\end{align}
We are interested in computing $\mathbb{E}[s]$ and $\mathbb{V}[s]$, which require integrating $s$ and $s^2$ against the distribution over $f'$. 
Instead of attempting to compute these quantities directly we sidestep the problem and first determine the probability distribution over the arc length integrand $(1+ (f')^2)^{1/2}$. 
\subsection{Integrand Distribution}

\begin{figure*}
\vspace{.3in}
 \includegraphics{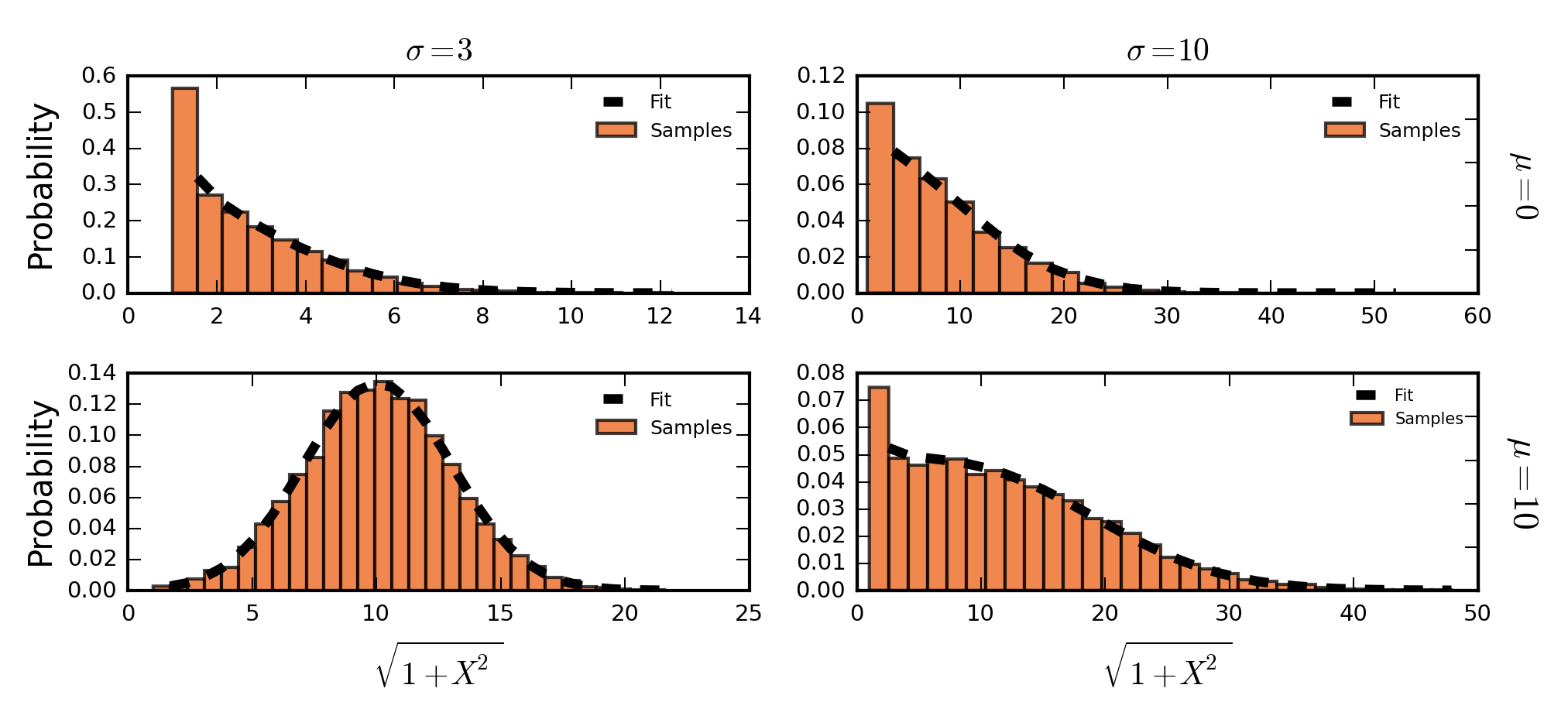}
 \caption{Histogram of samples from $\sqrt{1 + X^2}$, where $X\sim \mathcal{N}(\mu, \Sigma)$, overlaid with the corresponding distribution. We display the effects of varying $\mu$ and $\sigma$.} 
 \label{fig:1d_inc_disst}
\end{figure*}

We present a new method for deriving the mean and variance of the arc length of a one-dimensional \gls{gp} by first considering the transformation of a normal distribution variable under the non-linear transformation $g(x) = (1+x)^{1/2}$. 
Specifically, we can consider the distribution of a normally distributed random variable under the transformation $g$:
\begin{equation}
Y = g(X) = \sqrt{1+X^2}, \quad  X\sim \mathcal{N}(\mu, \sigma^2).
\end{equation}
We consider the more general case where $\mu \neq 0$. 
Intuitively, we expect our distribution for $Y$ to be a skewed Chi distribution. 
We are able to directly compute the probability density function for $Y$ by considering the cumulative distribution and using the standard rules for the transformation of probability functions: 
\begin{align}
 P(Y < y) & = F_{X}(\sqrt{y^2 - 1} + \mu) \notag \\
  & \quad- (1 - F_{X}(\sqrt{y^2-1}-\mu)),
\end{align}
where $F_{X}$ is the cumulative probability distribution of $X$; details in the Supplementary Material. 
The probability density function (pdf) of $Y$ is obtained by taking the derivative of $P(Y<y)$ with respect to $y$; further details in the Supplementary Material:
\begin{align}
 p_Y(y)  & = \frac{1}{\sqrt{2\pi}\sigma}\left[\exp\left(-\frac{(\sqrt{y^2 - 1} + \mu)^2}{2\sigma^2}\right)  \right. \notag \\ & \quad + \left.\exp\left(-\frac{(\sqrt{y^2 - 1} - \mu)^2}{2\sigma^2}\right) \right]\frac{y}{\sqrt{y^2-1}}.
\end{align}
This probability distribution is valid for $y>1$ and a straightforward calculation shows that $\int_{y \in Y}p_{Y}(y)\mathrm{d} y = 1$. 
Computation of the expectation of the integrand distribution can now be done in closed; the process is outlined in the Supplementary Material. The final expression is:
\begin{align}
  \mathbb{E}[y] & = \frac{1}{\sqrt{2\pi}\sigma}\exp\left(-\frac{\mu^2}{2\sigma^2}\right)\notag\\
  & \!\!\!\!\!\sum_{l=0}^{\infty}\frac{\Gamma\left(l+\frac{1}{2}\right)}{(2l)!}\left(\frac{\mu}{\sigma^2}\right)^{2l}U\left(l+\frac{1}{2},l+2,\frac{1}{2\sigma^2}\right)\!. \label{eqn:oned_inc_mean}
\end{align}
Here $\Gamma(n)$ is the gamma function and $U(a,b,z)$ is the confluent hypergeometric function of the second kind, defined by the integral expression:
\begin{align}
 U(a,b,z) & = \int^{\infty}_{0}\exp\left(-zt\right)t^{a-1}(1+t)^{b-a-1}\mathrm{d}t.
\end{align}
A similar process allows us to derive an exact expression for $\mathbb{E}_{p_Y(y)}[y^2]$ and hence $\mathbb{V}_{p_{Y}(y)}[y]$. Figure~\ref{fig:1d_inc_disst} shows draws of $g(X)$, overlaid with $p_{Y}(y)$ for a range of $\mu$ and $\sigma$. 

\subsection{Arc Length Statistics}
Having derived expressions for the arc length integrand distribution we are able to evaluate the moments of the arc length. 
Specifically, we consider a zero mean \gls{gp} with kernel $\mathrm{K}$:
\begin{align}
 f \sim \mathcal{N}(0,\mathrm{K})
\end{align}
The derivative process is a \gls{gp} \cite{Rasmussen2004} defined by:
\begin{align}
 f'\sim \mathcal{N}(0, \partial^2 \mathrm{K})
\end{align}
Taking the expectation of the arc length, noting that the integrand is non-negative, therefore by Fubini's Theorem \cite{Fubini1907} we can interchange the expectation and integral:
\begin{align}
    \mathbb{E}[s] & = \mathbb{E}\left[\int_{0}^{T}\sqrt{1 + (f')^2}\mathrm{d}t\right]\\
    & = \int_{0}^{T}\mathbb{E}\left[\sqrt{1 + (f')^2}\right]\mathrm{d}t.
\end{align}
The variance of $f'$ is given by $\sigma_{f'}^2 = R_{f'}(0)$. 
At each point along the integral the expectation of the integrand is the same, therefore we arrive at:
\begin{align}
    \mathbb{E}[s] & = \int_{0}^{T}\left[ \frac{1}{\sqrt{2\pi}\sigma_{f'}}\Gamma\left(\frac{1}{2}\right)U\left(\frac{1}{2} ,2 ,\frac{1}{2\sigma_{f'}^2}\right)  \right]\mathrm{d}t\\
    & =T\frac{1}{\sqrt{2\pi}\sigma}\Gamma\left(\frac{1}{2}\right)U\left(\frac{1}{2} ,2 ,\frac{1}{2\sigma_{f'}^2}\right),
\end{align}
where we have used the expectation of the integrand for the zero-mean case. 
Using identities related to the Confluent Hypergeometric we can rewrite the mean as:
\begin{align}
 \mathbb{E}[s] = \frac{ T \exp (1/4 \sigma_{f'}^2)}{2\sqrt{2\pi}\sigma_{f'}}\left[\mathrm{BF}_{0}\left(\frac{1}{4\sigma_{f'}^2}\right) + \mathrm{BF}_{1}\left(\frac{1}{4\sigma_{f'}^2}\right)\right],
\end{align}
where $\mathrm{BF}_i$ is the modified Bessel function of the second kind of order $i$. 
For a posterior distribution of the arc length, given data observations, we would use Eqn~\ref{eqn:oned_inc_mean} along with the the posterior derivative mean, $\mu_{f'} =  \frac{\partial m_*}{\partial x_*} $ and variance function of the posterior \gls{gp}, $\sigma_{f'}^2$ to compute the expected length;
\begin{align}
  \mathbb{E}[s] & = \sum_{l=0}^{\infty}\frac{\Gamma\left(l+\frac{1}{2}\right)}{(2l)!}\int^{T}_{0}\frac{1}{\sqrt{2\pi}\sigma_{f'}}\exp\left(-\frac{\mu_{f'}^2}{2\sigma_{f'}^2}\right)\notag\\
  & \!\!\!\!\!\left(\frac{\mu_{f'}}{\sigma_{f'}^2}\right)^{2l}U\left(l+\frac{1}{2},l+2,\frac{1}{2\sigma_{f'}^2}\right)\mathrm{d}t\!, \label{eqn:oned_inc_mean_post}
\end{align}
where $\mu_{f'} $ and $\sigma_{f'}$ depend on $t$. 
We have derived a closed form expression for the mean of the arc length of a one dimensional zero mean \gls{gp}, reproducing the original result from \cite{BARAKAT1970} whilst providing a way to compute the arc length mean of a \gls{gp} posterior distribution. 
The variance involves the computation of the second moment, a calculation involving the bi-variate form of the integrand distribution; we do not derive that in this paper. 
An alternate derivation is reported in \cite{BARAKAT1970}.

\subsubsection{Kernel Derivatives}
\begin{center}
\begin{figure}[h]
\vspace{.3in}
 \includegraphics{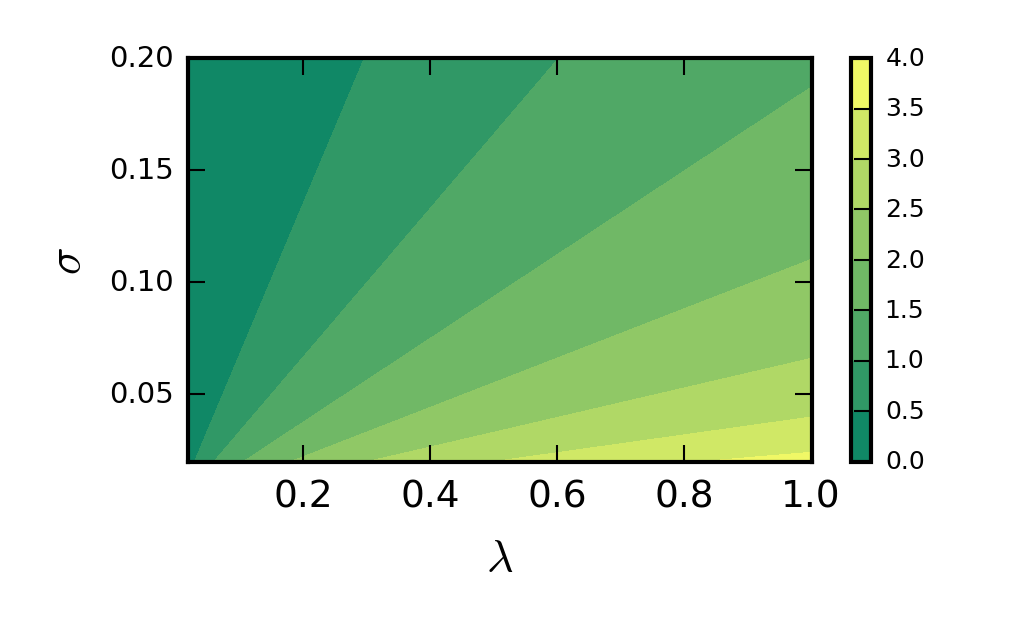}
 \caption{Values of the expected arc length (colour shading) for various values of the SE kernel parameters. The plot shows the heat map of the log of the arc length to show sufficient detail. The length is dominated by the input scale parameter.}
 \label{fig:se_heat_two}
\end{figure}
\end{center}

The value of the mean arc length is determined solely by the derivative variance, $\sigma_{f'}^2$. For stationary kernels, $k(x,x') = k(x-x')$, this equates to:
\begin{align}
\sigma_{f'}^2 & = \frac{\partial^2}{\partial x \partial x'}k(x-x')\bigg{|}_{x=x'}.
\end{align}
Table~\ref{tab:kernderivs} summarises a table of common kernels \cite{Rasmussen2004} and the variance of the effective length scale in terms of their hyperparameters. The effect of the choice of hyperparameters on the expected length is shown in Figure~\ref{fig:se_heat_two}.

\begin{table}[h]
\caption{Derivative process variance, $\sigma_{\dot{f}}^2$, in terms of kernel hyperparameters for a range of common kernels. In each case $\lambda^2$ is the output (signal) variance hyperparameter and $\sigma$ is the input dimension length scale hyperparameter.}
\label{tab:kernderivs}
\begin{center}
\begin{tabular}{p{2cm}p{1.5cm}p{1.5cm}p{1.5cm}}
\toprule
     Square Exponential & Mat\'{e}rn, $\nu= \frac{3}{2}$ & Mat\'{e}rn, $\nu= \frac{3}{2}$ &  Rational Quadratic\\
    \midrule 
    $\lambda^2/\sigma^2$ & $3\lambda^2/\mathbf{\sigma}^2$ & $5\lambda^2/3\sigma^2$ & $\lambda^2/\sigma^2$ \\
    \bottomrule
\end{tabular}
\end{center}
\end{table}


\section{MULTI-DIMENSIONAL ARC LENGTH}
\label{sec:general_case}
In this section we present the first treatment of the arc length of a \gls{gp} in more than one output dimension. We present an approximation to the arc length integrand distribution and use this to compute the moments of the arc length. For the vector case, we now consider a vector \gls{gp} and its corresponding derivative process:
\begin{align}
 \mathbf{f} \sim \mathrm{GP}(0, \mathbf{K}), \quad \mathbf{f}' \sim \mathrm{GP}(0, \partial^2\mathbf{K}),
\end{align}
where $\mathbf{K} = \mathrm{B}\otimes \mathbf{k}$, with a coregionalised matrix $\mathrm{B}$ and a stationary kernel $\mathbf{k}$. The arc length for the vector case is given by:
\begin{align}
 s = \int_{a}^{b}|\mathbf{f}'|\mathrm{d}t.
\end{align}
As we did in the one-dimensional case we first consider the distribution of the arc length integrand $|\mathbf{f}'|$ and then use this to derive the moments of the arc length itself. 

\subsection{Integrand Distribution}

We are interested in the distribution over the arc length. 
Ultimately we are interested in $\mathbb{R}^3$, however, the theory we present is valid for any $\mathbb{R}^n$. 
We consider the random variable $\mathrm{W}$, defined by:
\begin{align}
 \mathrm{W} & = |\mathbf{x}| = (\mathbf{x}^T\mathbf{x})^{1/2}  = \sqrt{\sum^n_{i}\mathbf{x}_{i}^2} \\
  \mathbf{x} & \sim \mathcal{N}(\mu, \Sigma),
\end{align} 
with $\mathbf{x}, \mu \in \mathbb{R}^n$ and $\Sigma \in \mathbb{R}^{n\times n}$ is a full-rank covariance matrix. 
This is the square root of the sum of squares of correlated normal variables. 
It is well know that the sum of squares of independent identically distributed normal variables is Chi-squared distributed and that the corresponding square root is Chi distributed \cite{Johnson2}. 
At first glance it seems that we should easily be able to identify this transformed distribution, however, the full-covariance between the elements of $\mathbf{x}$ hinder the derivation of a straightforward distribution.

Substantial work has been done on the distribution of quadratic forms, $Q(\mathbf{x})= \mathbf{x}^TA\mathbf{x}$ \cite{Mathai1992}, where $\mathbf{x}$ is an $n \times 1$ normal vector defined previously and $A$ is a symmetric $n \times n$ matrix. It is possible to write:
\begin{align}
 Q(\mathbf{x}) & = \mathbf{x}^TA\mathbf{x} = \sum_{i}^{n} \lambda_i(U_i + b_i)^2,  \label{eqn:quadratic_form_expan}
\end{align}
where the $U_i$ are $ $ i.i.d. normal variables with zero mean and unit variance, the $\lambda_i$ are the eigenvalues of $\Sigma$ and $b_i$ is the $i$th component of $b = P^T\Sigma^{\frac{1}{2}}\mu$, with $P$ a matrix that diagonalises $\Sigma^{\frac{1}{2}}A\Sigma^{\frac{1}{2}}$ . 

Observing the summation of the quadratic form in Eqn \ref{eqn:quadratic_form_expan}, we see that our distribution is a weighted sum of Chi-squared variables. Unfortunately, there exists no simple closed-form solution for this distribution, however, it is possible to express this distribution via power-series of Laguerre polynomials and some approximations have been used \cite{Mathai1992}.

\begin{figure*}
\vspace{.3in}
 \includegraphics[trim={0 0 0 10}]{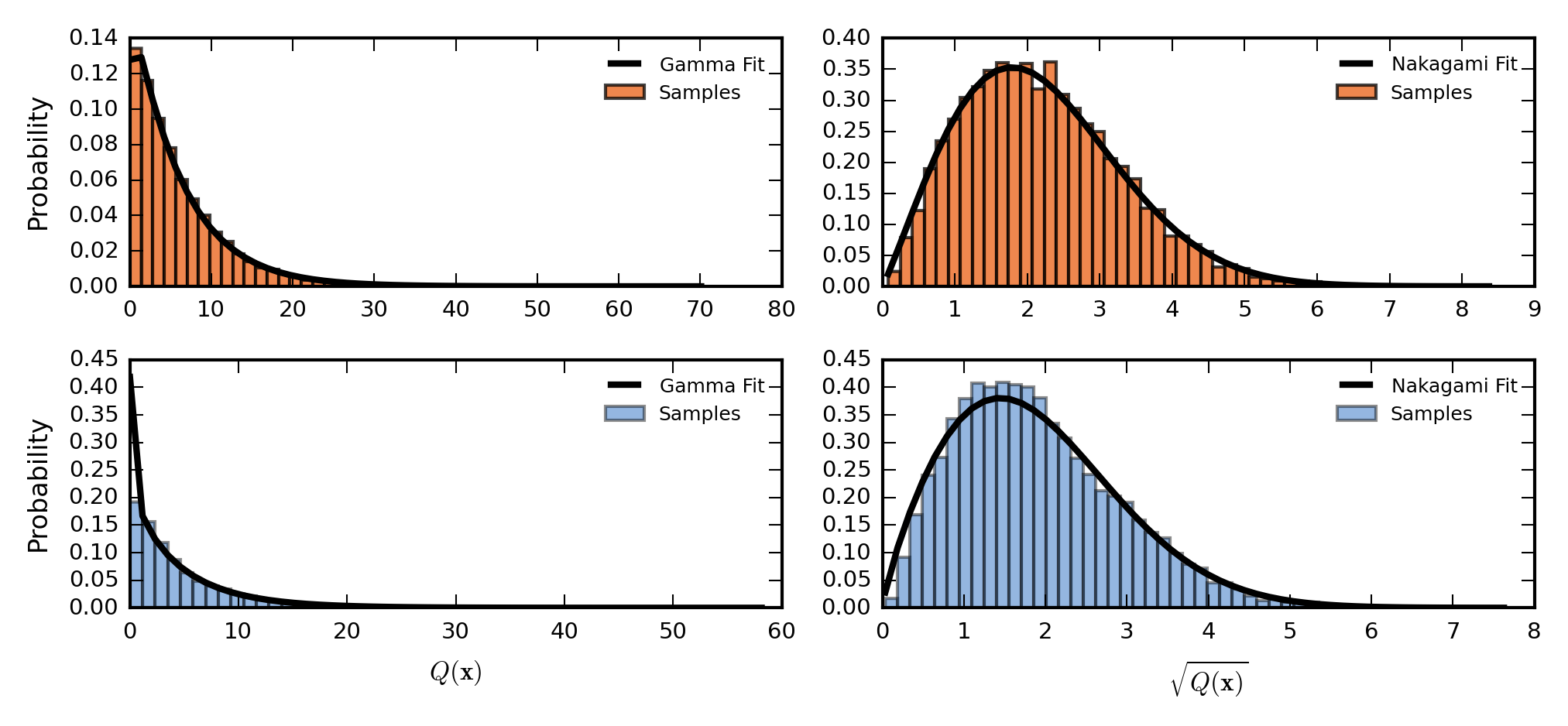}
 \caption{Samples from $Q(\mathbf{x})$ and $\sqrt{Q(\mathbf{x})}$ overlaid with the approximated gamma and Nakagami distributions. $\mathbf{x} \sim (0, \Sigma)$ in the top row, and $\mathbf{x} \sim (\mu, \Sigma)$ in the bottom row with $\mu$ and $\Sigma$ randomly generated. Similar plots are obtained for different values of $\mu$ and $\Sigma$. The gamma and Nakagami distributions provide a reasonable approximation to the shape of the distribution, whilst capturing the true mean and variance. }
 \label{fig:3d_mom_approx}
\end{figure*}

We note that a Chi-squared distribution is a gamma distributed variable for the case where the shape parameter is $v/2$ and the scale factor is 2. 
Therefore we will approximate $Q(\mathbf{x}) = \mathbf{x}^T\mathbf{x}$ with a single gamma random variable by moment matching the first two moments. 
The mean and variance of $Q(\mathbf{x})$ are given by:
\begin{align}
 \mathbb{E}[Q(\mathbf{x})] & = \text{tr}(\Sigma ) + \mu^T\mu, \\
  \mathbb{V}[Q(\mathbf{x})] & = 2\text{tr}(\Sigma\Sigma) + 4 \mu^T\Sigma \mu,
\end{align}
where $\text{tr}()$ denotes that trace of a matrix. The pdf of a gamma distribution with shape $k_G$ and scale $\theta_G$ is given by 
\begin{align}
 p_{G}(x: k_G, \theta_G) = \frac{x^{k_{G}-1}\exp\left(-\frac{x}{\theta_G}\right)}{\theta_{G}^{k_G} \Gamma(k_G)}.
\end{align}
The first two moments are:
\begin{align}
  \mu_G & = k_G\theta_G, \quad  \sigma_G^2 = k_G\theta_G^2.
\end{align}
Solving for $k_G$ and $\theta_G$:
\begin{equation}
 k_G  = \frac{\mu_G^2}{\sigma_G^2}, \quad \theta_G  = \frac{\sigma_G^2}{\mu_G}.
\end{equation}
Equating moments, we set $\mu_G = \mathbb{E}[Q(\mathbf{x})] $  and $\sigma_G^2 = \mathbb{V}[Q(\mathbf{x})]$. Thus, $Q$ is approximated as a gamma random variable and we write, $Q(\mathbf{x}) \sim \text{Gamma}(k_G, \theta_G)$. 
\\\\
Now we are in a position to consider the quantity $\sqrt{Q}$. 
Here we use that fact that if a random variable $Q \sim \text{Gamma}(k_G,\theta_G)$, then the random variable $\mathrm{W} = \sqrt{Q}$ is a Nakagami random variable $\mathrm{W} \sim \text{Nagakami}(m, \Omega)$, with parameters given by $m = k_G$ and $\Omega = k_G\theta_G$. 
The nagakami distribution \cite{Hoffman2013} is:
\begin{align}
 p_{\mathrm{Nak}}(x; m, \theta) = \frac{2m^m}{\Gamma(m)\Omega^m}x^{2m-1}\exp\left(-\frac{m}{\Omega}x^2\right).
\end{align}
Using the value for $k$ and $\theta$ obtained via our moment matched approximation and transforming to the Nakagami distribution we say $\sqrt{Q}$ is approximated as a Nakagami distribution with parameters:
\begin{equation}
 m = \frac{\mu_{G}^2}{\sigma_{G}^2}, \quad \Omega = \mu_{G}.
\end{equation}
In terms of our original distribution $\mathbf{x} \sim \mathcal{N}(\mu,\Sigma)$, we therefore have $\mathrm{W} = \sqrt{\mathbf{x}^T\mathbf{x}} \sim \text{Nakagami}(m,\Omega)$, with:
\begin{equation}
 m = \frac{[ \text{tr}(\Sigma ) + \mu^T\mu]^2}{2\text{tr}(\Sigma\Sigma) + 4 \mu^T\Sigma \mu}, \quad \Omega = \text{tr}(\Sigma ) + \mu^T\mu.
\end{equation}
The mean and variance are:
\begin{align}
 \mathbb{E}[\mathrm{W} ] & = \frac{\Gamma(m + \frac{1}{2})}{\Gamma(m)}\left(\frac{\Omega}{m}\right)^\frac{1}{2}\\
\mathbb{V}[\mathrm{W}] & = \Omega\left( 1 - \frac{1}{m}\left( \frac{\Gamma(m+\frac{1}{2})}{\Gamma(m)} \right)^2 \right).
\end{align}
The method we have used to derive the distribution of the arc length integrand is summarised in Eqn \ref{eqn:flow}:
\begin{align}
 \mathcal{N}(\mu, \Sigma) \underset{\text{Approximate}}{\overset{Q}{\rightarrow}} \text{Gamma}(k_G, \theta_G) \underset{\text{Exact}}{\overset{\sqrt{Q}}{\rightarrow}} \text{Nakagami}(m, \Omega)
 \label{eqn:flow}
\end{align}
Numerical samples of $Q(\mathbf{x})$ and $\sqrt{Q(\mathbf{x})}$ and the pdf of the corresponding gamma and Nakagami distributions are show in Figure~\ref{fig:3d_mom_approx} for $d=3$. 
The approximated distributions show a reasonable approximation for a range of $\mu$ and $\Sigma$.

The quadratic form approximated to the gamma distribution is exact when all the eigenvalues of the covariance are identical, in that case we have only a single gamma random variable. 


\subsection{Arc Length Statistics}

We are now in a position to consider the arc length directly. 
Taking the expectation of the arc length, recalling that expectation is a linear operator and using Fubini's theroem:
\begin{align}
    \mathbb{E}[s]
    & = \int_{0}^{T}\mathbb{E}\left[(\mathbf{f}'^T\mathbf{f}')^{\frac{1}{2}}\right]\mathrm{d}t.
\end{align}
Recalling the form of our kernel as $\mathbf{K}(x,x') = \mathrm{B}\otimes k(x,x')$, the infinitesimal distribution of $\mathbf{f}'$ is constant with respect to $t$ with covariance given by: 
\begin{align}
\Sigma_{f'} = \mathrm{B}\otimes \frac{\partial^2}{\partial x \partial x'} k(x,x')\bigg{|}_{x=x'} = \mathrm{B}~\sigma_{f'}^2.
\end{align}
Therefore the expected length of the arc length is:
\begin{align}
 \mathbb{E}[s] 
 & \approx T\frac{\Gamma(m_{f'}+\frac{1}{2})}{\Gamma(m_{f'})}\left(\frac{\Omega_{f'}}{m_{f'}}\right)^{\frac{1}{2}},\label{eqn:first_moment}
\end{align}
with,
\begin{align}
 m_{f'} = \frac{[\text{tr}(\Sigma_{f'})]^2}{2\text{tr}(\Sigma_{f'}\Sigma_{f'})}, \quad \Omega_{f'} = \text{tr}(\Sigma_{f'}),
\end{align}
where we have used the Nakagami approximation to the arc length integrand to evaluate the mean. 
As in the one-dimensional case, the expected length of the \gls{gp} is determined solely by the choice of kernel and the length of the interval. 
The calculation of the variance requires the second moment:
\begin{align}
 \mathbb{E}[s^2]
 & = \int\int\mathbb{E}\left[|\mathbf{f}'_{t_1}|~|\mathbf{f}'_{t_2}|\right]\mathrm{d}t_1\mathrm{d}t_2.
\end{align}
Making use of the Nakagami approximation to our integrand we need the mixed moment of two correlated Nakagami variables. 
Let us write $|\mathbf{f}'_{t_1}| \approx \mathrm{W}_1$, $|\mathbf{f}'_{t_2}| \approx \mathrm{W}_2$, with $\mathrm{W}_1 \sim \text{Nakagami}(m_{f'}, \Omega_{f'})$ and $\mathrm{W}_2 \sim \text{Nakagami}(m_{f'}, \Omega_{f'})$. 
The mixed moments of two correlated Nakagami variables with the same parameters is given by \cite{Reig2002}:
\begin{align}
 \mathbb{E}[\mathrm{W}_1^n\mathrm{W}_2^l] & = \frac{\Omega}{m}\frac{[\Gamma(m+n/2)]^2}{[\Gamma(m)]^2}{}_{2}F_{1}\left(-\frac{n}{2},-\frac{l}{2},m: \rho(\tau) \right),
\end{align}
where $\rho(\tau)$ is the correlation between the gamma variables that the Nakagami distribution was derived from and ${}_2F_{1}$ is the hypergeometric function:
\begin{align}
 {}_{2}F_{1}(a,b,c:z) & = \sum_{n=0}^{\infty}\frac{(a)_n(b)_n}{(c)_n}\frac{z^n}{n!}
\end{align}
with the Pochhammer $(q)_n$ symbol defined as $(q)_n = q(q+1)\dots(q+n-1) $ and $(q)_0= 1$. The second moment can now be expressed as a power series in $\rho$:
\begin{align} 
 \mathbb{E}[s^2] 
 & = \frac{\Omega}{m}\frac{[\Gamma(m+1/2)]^2}{[\Gamma(m)]^2}\sum_{n=0}^{\infty}\frac{\left(-\frac{1}{2}\right)_n\left(-\frac{1}{2}\right)_n}{(m)_n}\frac{1}{n!}\notag \\ &\qquad\int_0^T\int_0^T\rho(t_1-t_2)^n\mathrm{d}t_1\mathrm{d}t_2. \label{eqn:second_moment}
\end{align}
We derive the correlation function:
\begin{align}
 \rho(t-t') & = \left[\frac{\partial^2}{\partial t \partial t'} k(t,t')\right]^2 \frac{1}{\sigma_{f'}^4}.
\end{align}
Eqn~\ref{eqn:second_moment} can be solved numerically (noting that the two-dimensional integral is readily tackled using traditional methods of quadrature) and the variance is then computed by $\mathbb{V}[s] = \mathbb{E}[s^2] - \mathbb{E}[s]^2$.

\subsection{Arc Length Posterior}
The moments of the arc length of a \gls{gp} posterior follow a similar derivation. 
The posterior mean of the arc length is:
\begin{align}
\mathbb{E}[s] & \approx \int_{0}^{T}\frac{\Gamma(m_{f'}+\frac{1}{2})}{\Gamma(m_{f'})}\left(\frac{\Omega_{f'}}{m_{f'}}\right)^{\frac{1}{2}}\mathrm{d}t,
\label{eqn:arc_mean_post}
\end{align}
where $m_{f'}$ and $\Omega_{f'}$ are the Nakagami parameters which now depend on the mean and covariance functions of the \gls{gp} posterior, which themselves are functions of $t$. This non tractable expression now requires an integration (which, again, can be efficiently approximated with quadrature). The posterior second moment is given by:
\begin{align}
\mathbb{E}[s^2] & \approx \sum_{n=0}^{\infty}\frac{\left(-\frac{1}{2}\right)_n(-\frac{1}{2})_n}{n!}\int_0^T\int_0^T\left(\frac{\Omega_1}{m_1}\right)^{\frac{1}{2}}\left(\frac{\Omega_2}{m_2}\right)^{\frac{1}{2}}\notag\\ &\quad\frac{\Gamma(m_1 + 1/2)\Gamma(m_2+1/2)}{\Gamma(m_1)\Gamma(m_2)(m_{2})_n}\rho(|t_1-t_2|)^n\mathrm{d}t_1\mathrm{d}t_2,
\label{eqn:arc_var_post}
\end{align}
where $m_i$ and $\Omega_i$ again depend on the mean and covariance functions of the \gls{gp} posterior and are evaluated at $t_i$.

\section{SIMULATIONS}
\label{sec:numerical}

In this section we generate samples from our \gls{gp} prior and compute the arc length, focusing on the vector case. 
We show the effect of the kernel choice and show the fidelity of our theoretical results. 
To generate our curves we specify a zero mean \gls{gp} kernel, $\mathrm{K} = \mathrm{B}\otimes k(t,t')$, with fixed $\mathrm{B}$ and we use the Mat\'{e}rn Kernel with $\nu = 3/2$, which we call the M32 kernel: 
\begin{align}
 k(t,t') & = \lambda^2\left( 1 + \frac{\sqrt{3}||t-t'||}{\sigma} \right)\exp\left(-\frac{\sqrt{3}||t-t'||}{\sigma}\right).
\end{align}
We draw a sample $f_i = (x_i, y_i, z_i)$ evaluated at evenly spaced $t$. 
The arc length of the \gls{gp} draw is then computed numerically.

\begin{figure}
\vspace{.3in}
 \includegraphics[trim={0 0 0 10}]{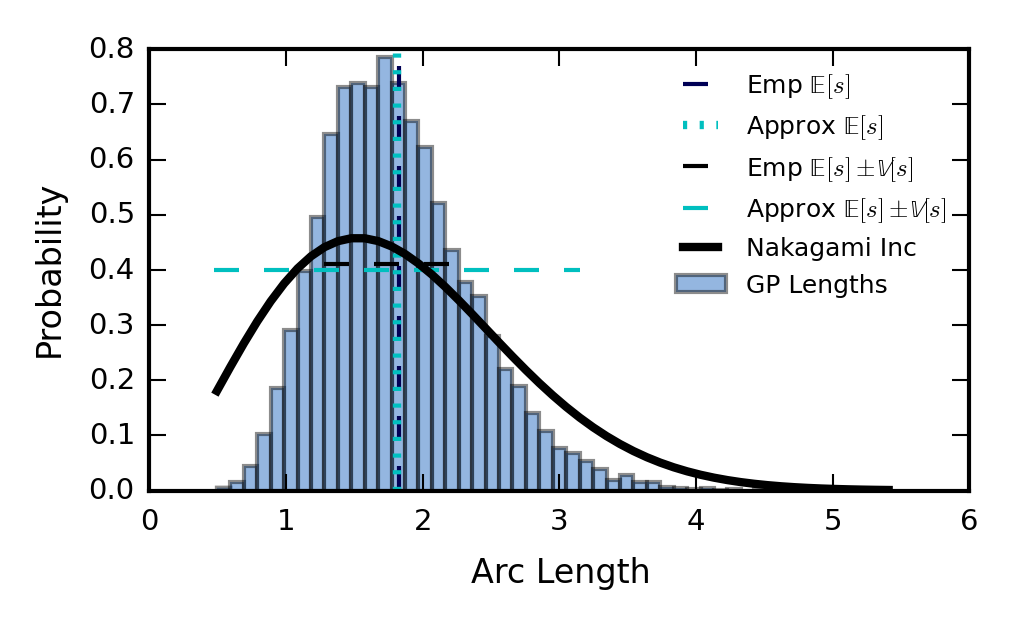}

 \caption{Histogram of GP Lengths. The theoretical and empirical mean are shown and the corresponding variance. The Nakagami distribution of the integrand is also shown. We can see the integral over integrands has the effect of shrinking the variance relative to a single integrand.}
 \label{fig:3d_samples_arc}
\end{figure}

Unit variance and length scale parameters are chosen and the arc length is computed over the interval $t = [0,1]$. 
Figure~\ref{fig:3d_samples_arc} shows the sample lengths, the theoretical mean and variance, and the Nakagami distribution of a single arc length integrand. 
Our theoretical results are close to the numerically generated values. 
The plot of the Nakagami distribution demonstrates the wide variance of an individual arc length integrand with respect to the overall variance. 
We see that the integration over the input domain has a sort of `shrinking' effect on overall variance when compared to the individual variance.
No estimation methods are required to calculate the arc length statistics. 
Our approximated equations are closed form (for the mean) and a quadrature problem (for the variance). 






 
\section{CONCLUSION}
In this paper we derive the moments of a vector valued \gls{gp}. 
To the best of the authors' knowledge, this is the first treatment of the arc length in more than one dimension. 
The increment distribution was approximated via its moments to a Nakagami distribution which provide a closed form for the mean, Eqn \ref{eqn:first_moment} and an expression for the second moment, Eqn \ref{eqn:second_moment}, of the arc length.
Importantly, we are also able to derive the first, Eqn \ref{eqn:arc_mean_post}, and second, Eqn \ref{eqn:arc_var_post}, moment of the arc length of a \gls{gp} posterior, conditioned on observations of the function.

The moments were shown to depend on the choice of kernel, the hyperparameters and the length of the interval. Numerical experiments confirmed the fidelity of our approximation to the arc length integrand and the arc length moments. We also provide a visual understanding of the distribution. 

We see knowledge of the arc length as a valuable tool which will allow us to encode more information into our prior over kernel choices. The explicit relation between the arc length moments and the kernel hyperparameters allow us to use prior information to better initialize and constrain our models, in particular in cases where lengths correspond to interpretable quantities, such as a path trajectory.  


Potential avenues of future research include analysis of the non-stationarity of curves, curve minimisations problems, and generating curves of a given length. Furthermore, we see potential application in Bayesian optimization, as a path planning tool and for constructing interpretable features from functional data.

\subsubsection*{Acknowledgements}

AT and MO are grateful for the support of funding from the Korea Institute of Energy Technology Evaluation and Planning (KETEP).

The authors are grateful for initial conversations with Tom Gunter who highlighted the gap in the literature.



\bibliographystyle{siam} 

\bibliography{library.bib}

\newpage
\onecolumn
\section*{Supplementary Material}

\subsection*{Density of $Y = \sqrt{1+X^2}$}
Cumulative distribution of $Y$:
\begin{align}
 P(Y < y) & = P(|X-\mu| < \sqrt{y^2-1})\\
 & = P(-\sqrt{y^2-1} < X- \mu < \sqrt{y^2-1})\\
 & = P(-\sqrt{y^2-1} + \mu < X < \sqrt{y^2-1} +\mu) \\
 & = F_{X}(\sqrt{y^2 - 1} + \mu) - (1 - F_{X}(\sqrt{y^2-1}-\mu)).
\end{align}
Probability density of $Y$:
\begin{align}
 p_Y(y) & = \frac{\mathrm{d}}{\mathrm{d} y}P(Y<y)\\
 & = \frac{\mathrm{d}}{\mathrm{d} y}\left[F_{X}(\sqrt{y^2 - 1} + \mu)- (1 - F_{X}(\sqrt{y^2-1}-\mu))\right]\\
 & = \frac{1}{\sqrt{2\pi}\sigma}\left[\exp\left(-\frac{(\sqrt{y^2 - 1} + \mu)^2}{2\sigma^2}\right)  \right. \notag  + \left.\exp\left(-\frac{(\sqrt{y^2 - 1} - \mu)^2}{2\sigma^2}\right) \right]\frac{y}{\sqrt{y^2-1}}.
\end{align}
Mean of $Y$:
\begin{align}
\mathbb{E}_{p_Y(y)}[y]& = \int_{1}^{\infty}y~{p}_{Y}(y)\mathrm{d}y \\
 & = \frac{1}{\sqrt{2\pi}\sigma}\int_{1}^{\infty}\left[\exp\left(-\frac{(\sqrt{y^2 - 1} + \mu)^2}{2\sigma^2}\right) + \exp\left(-\frac{(\sqrt{y^2 - 1} - \mu)^2}{2\sigma^2}\right) \right]y^2(y^2 - 1)^{-\frac{1}{2}}\mathrm{d}y.
\end{align}
At first glance this looks to be an intractable integral, however, with the change of variables $y = (x^2 + 1)^{1/2}$ and by expanding the exponential cross terms we arrive at:
\begin{align}
  \mathbb{E}[y] & = \frac{1}{\sqrt{2\pi}\sigma}\exp\left(-\frac{\mu^2}{2\sigma^2}\right)\sum_{l=0}^{\infty}\frac{\Gamma\left(l+\frac{1}{2}\right)}{(2l)!}\left(\frac{\mu}{\sigma^2}\right)^{2l}U\left(l+\frac{1}{2},l+2,\frac{1}{2\sigma^2}\right)\!. \label{eqn:oned_inc_mean}
\end{align}





\end{document}